\documentclass[]{interact}

\usepackage{epstopdf}% To incorporate .eps illustrations using PDFLaTeX, etc.
\usepackage{subfigure}% Support for small, `sub' figures and tables
\usepackage{float}
\usepackage{natbib}% Citation support using natbib.sty
\bibpunct[, ]{(}{)}{;}{a}{}{,}% Citation support using natbib.sty
\renewcommand\bibfont{\fontsize{10}{12}\selectfont}% Bibliography support using natbib.sty

\usepackage{xcolor}

\newcommand{\rthird}{\textcolor{blue}}

\begin{document}

\title{Physics-based Simulation Ontology: an ontology to support modeling and reuse of data for physics-based simulation}

\author{\name{Hyunmin Cheong, Adrian Butscher \thanks{hyunmin.cheong@autodesk.com, adrian.butscher@autodesk.com}}
\affil{Autodesk Research, 661 University Ave, Toronto, ON, Canada M5G 1M1}}

\maketitle

\begin{abstract}
The current work presents an ontology developed for physics-based simulation in engineering design, called Physics-based Simulation Ontology (PSO). The purpose of the ontology is to assist in modeling the physical phenomenon of interest in a veridical manner, while capturing the necessary and reusable information for physics-based simulation solvers. The development involved extending an existing upper ontology, Basic Formal Ontology, to define lower-level terms of PSO. PSO has two parts -- PSO-Physics, which consists of terms and relations used to model physical phenomena based on the perspective of classical mechanics involving partial differential equations, and PSO-Sim, which consists of terms used to represent the information artifacts that are about the physical phenomena modeled with PSO-Physics. The former terms are used to model the physical phenomenon of interest independent of solver-specific interpretations, which can be reused across different solvers, while the latter terms are used to instantiate solver-specific input data. A case study involving two simulation solvers was conducted to demonstrate this capability of PSO. Discussion around the benefits and limitations of using BFO for the current work is also provided, which should be valuable for any future work that extends an existing upper ontology to develop ontologies for engineering applications.
\end{abstract}

\begin{keywords}
Ontologies; knowledge representation; CAE; CAD; physics-based simulation
\end{keywords}

%=================================================

\section{Introduction}
\label{intro}

Physics-based simulation has become an essential part of engineering design. In many industries, computer-aided engineering (CAE) tools have been widely used to analyze the physical behavior of artifacts to be created, mainly to validate their requirements.  Simulation is also the core of design optimization or generative design, in which software computes for optimal designs based on the evaluation of their physical behavior and qualities. Moreover, simulation is necessary for virtual commissioning via digital twins, which are intended to emulate the corresponding artifacts and systems in physical reality. These trends indicate the increased use and importance of physics-based simulation in engineering going forward.

In this context, important challenges regarding the data used by simulation tools must be addressed. First, the data representing some physical phenomena to be simulated must be modeled in a veridical and consistent manner. There needs a framework that can assist the user to accurately model the physical phenomena of interest while capturing the necessary information that is required by simulation solvers.

Currently, modeling a simulation problem typically requires the user to adopt application-specific interpretations, a form of ``conceptualization" \citep{gruber1995toward} of reality. There is no assurance that the data modeled based on such interpretations actually correspond to the physical phenomenon to be simulated. Furthermore, different conceptualizations can create inconsistent views of the same physical phenomenon that fundamentally cannot be shared across different applications. In contrast, one could develop an ontology to establish the common viewpoint of reality \citep{guarino1998formal, smith2004beyond}. Then, any data modeled using the ontology could share the veridical and consistent viewpoint. Such models could then be translated as input data to specific solvers, during which application-specific interpretations can be made. With this approach, the differences between solvers due to their own ontological interpretations and numerical implementations can be isolated from the differences in user modeling, and the results from different solvers can be objectively compared.

Another significant challenge that needs to be addressed is to support the sharing and reuse of data across different solvers. For example, in generative design or virtual commissioning, multiple solvers might be used to simulate different aspects of the physical behavior anticipated for the object of interest. During this process, it would be ideal to reuse as much data as possible between different solvers.

A well-known problem that prevents the reuse of data is the lack of a common vocabulary that is shared between applications. Different solvers can refer to the same referent entity using different names. For example, input data to different solvers may feature ``Young's modulus", ``elastic modulus", ``modulus of elasticity", or ``E", which all refer to the same material property. Ideally, the values defined for such data items, when already defined for a particular solver, should be reused again for another solver. Ontologies have been well-known to address this type of challenge by serving as a controlled, reference vocabulary \citep{gruber1995toward, uschold1996ontologies}

At the same time, some parts of the data modeled for one simulation solver may be reusable for another solver while other parts may not. For example, one solver might require a mesh file for discrete representation of the simulated object, while another solver might use a voxel file. In such scenario, the specific discrete representation used by each solver cannot be shared, but sharing the overall shape of the design object, which may be represented using an application-neutral format such as STEP, would be useful. Other examples of data that can be reused include material properties and the duration of a physical phenomenon, while data such as simulation time steps (i.e., discretization of the duration) should be defined for each solver. A framework to distinguish these two types of data would facilitate the reuse of data across solvers.

%--------------------------------------------------

\subsection{Research questions}

To address the challenges identified above, the current work takes an ontological approach with the following questions in mind:

\begin{itemize}
\item Can an ontology serve as the framework to model some physical phenomenon of interest in a veridical and consistent manner while capturing the necessary information required for simulation solvers?
\item Can an ontology help the reuse and sharing of data, particularly by distinguishing the types of information modeled that can be reused across different solvers versus those that need to be redefined?
\end{itemize}

To answer these questions, the current work extends an existing upper ontology to develop an ontology called Physics-based Simulation Ontology (PSO). The following section explains the approach.

\subsection{Proposed approach}

The first research question prompted us to adopt an upper ontology developed with explicit commitment to ontological realism, namely Basic Formal Ontology 2.0 (BFO) \citep{smith2015bfo}, as the basis for developing PSO. The reasons for this choice are as follows.

Ontological realism aims at representing reality as it exists, mainly based on empirical findings \citep{smith2010ontological}. Subsequently, the assertions made in the ontology should always have truth correspondence in reality \citep{mulligan1984truth}. This view aligns well with the goal of developing an ontology that can model the physical phenomena of interest as veridically as possible.

One may question why ontological realism is relevant to physics-based simulation that operates in digital environments. That is because good physics-based simulation must conform to the constraints of physical reality for its results to be any meaningful \citep{turnitsa2010ontology}. Also, input data to simulation often come from reality, e.g., force measurements, and output data from simulation are compared against observations from reality, e.g., to validate design specifications.

To develop an ontology for physics-based simulation, we must choose a particular perspective of reality that is commonly assumed by most simulation solvers. That perspective is based on using classical mechanics to explain physical behaviors occurring at macroscopic levels, in contrast to other perspectives found in physics such as those based on relativistic or quantum mechanics. More precisely, we take the view that the laws of physics can be described using partial differential equations and the physical behaviors of objects can be predicted by solving boundary value problems involving those equations.

This commitment to a particular perspective of reality is in line with the principles of ontological realism. As emphasized in \cite{arp2015building}, ontological realism embraces \emph{perspectivism}, which accepts that multiple perspectives can be valid as long as each of them is veridical. In science, multiple competing theories may exist to explain reality at different levels of granularity, e.g., classical vs. quantum mechanics. Yet, this does not preclude a realist ontology to choose one particular perspective that is most useful for its application.

The choice of BFO also helps us answer the second research question. BFO, while primarily developed to represent physical reality, has developed a technique to deal with \emph{information artifacts}, which are representational entities that are about some other things \citep{ceusters2015aboutness}. This distinction of information artifacts helps us demarcate the information content entities that are solver-specific from the model of physics entities, which is ideally solver-neutral.

Finally, the current work strives for reusing existing ontologies. Our goal is not to unnecessarily reinvent ontologies and ignore the substantial work done in the field. Instead of creating a new ontology from scratch, we have extended a well-established ontology and reused theories from other existing work. 

\begin{figure}[!b]
\begin{center}
\includegraphics[width=\linewidth]{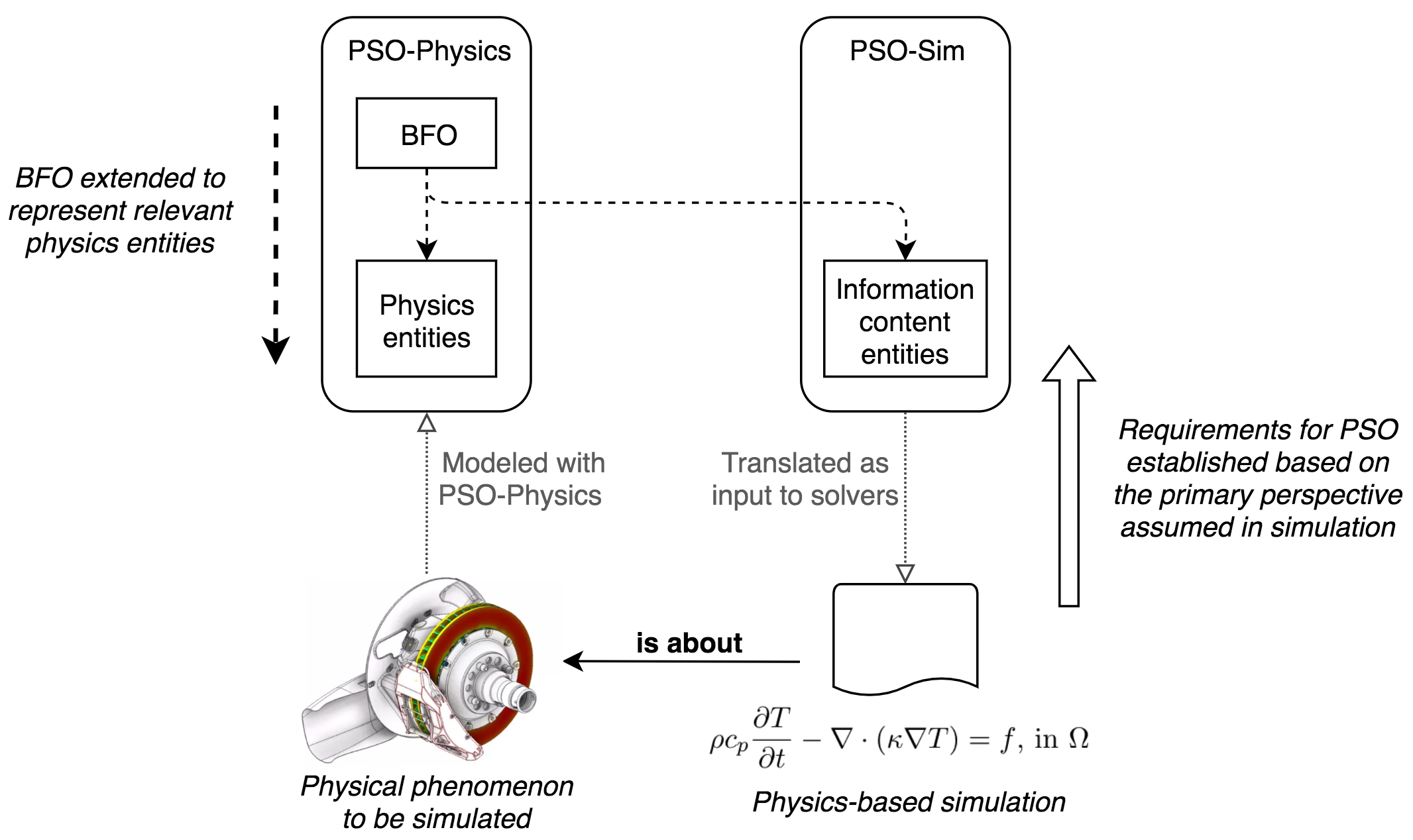}
\caption{A chosen approach to develop and use PSO.}
\label{fig1}
\end{center}
\end{figure}

Our overall approach, depicted in Figure~\ref{fig1}, can be summarized as follows:

\begin{itemize}
 \setlength\itemsep{1em}
 
  \item First, the scope and requirements of PSO are identified by examining the primary perspective assumed by the ontology. We take the perspective of classical mechanics, with the assumption that the behavior of three-dimensional objects participating in physical processes at macroscopic levels can be described using partial differential equations. In fact, this is the perspective assumed by the majority of the physics-based simulation solvers used in CAE.
  
  \item Next, we take BFO \citep{smith2015bfo} as the backbone and attempt to extend it to include categories representing the physics entities that are relevant to the requirements identified. We consider existing ontological theories, such as mereotopology, material constitution, 3D vs. 4D perspectives, etc., and take the most applicable theories to create the extensions.
  
  \item Information content entities (ICE) are created to represent input data for simulation solvers. ICEs are things such as documents, digital files, or images that are about some other entity \citep{ceusters2015aboutness}. For our purpose, ICEs can be thought as solver-specific data categories that refer to the physics entities identified in the above step. Example ICE terms include \emph{domain}, \emph{boundary condition}, \emph{time step}, etc. The current paper identifies a few terms as examples that are widely accepted by most solvers.

\end{itemize}

The above approach results in an ontology consisting of a group of terms used to model the physical phenomenon to be simulated and another group of terms that refer to the former terms to define input data for a particular simulation solver. Hereinafter, the first group of terms will be referred as PSO-Physics, while the latter group of terms will be referred as PSO-Sim.

%--------------------------------------------------

\subsection{Contributions}

Our first main contribution is the ontology developed, Physics-based Simulation Ontology (PSO). PSO is the first of its kind developed to assist engineers in modeling physical phenomena of interest in a veridical and consistent manner, which can subsequently be used as valid input to simulation solvers. The key feature of the ontology is the distinction of the types of information modeled, the objective representation of physical phenomena versus the solver-specific interpretations of those phenomena, the former of which can be reused across different solvers. Such an ontology and its framework are crucial to resolving the interoperability issues that are commonly observed among CAD/CAE systems. It is also a necessary step toward supporting semantic reasoning of physics-based simulation models such as for data validation \citep{cheong2019translating}, qualitative analysis of physics \citep{bahar2019}, and solver recommendation.

In addition, the current work demonstrates the validity and effectiveness of the realism-based ontological engineering approach in developing an ontology for engineering design. Namely, the current work shows how an existing upper ontology, BFO, can be extended to create a more domain-specific ontology and its distinction of physics entities versus information entities serves as an effective framework for ontology development. It also shows that ontological realism can be useful and relevant for the domain of engineering design, and not just in natural sciences in which the success has been demonstrated \citep{smith2007obo}.

\subsection{Outline of the paper}

The rest of the paper is organized as follows. A literature review is presented, followed by the scope and requirements for PSO. Then, PSO-Physics and PSO-Sim are presented. A case study involving the use of PSO to model an example problem and provide input data to two different simulation solvers is described. Finally, the paper ends with discussion and conclusions.

%=================================================

\section{Related work}

First discussed are the benefits of using an upper ontology, a review of existing upper ontologies, and justification for choosing BFO as the upper ontology for our work. Then, various applications of ontologies for CAE / CAD (computer-aided design) / PLM (project life-cycle management) are reviewed. Lastly, other work related to creating ontologies for physics and simulation is presented.

%--------------------------------------------------

\subsection{Upper ontologies}

Upper, or top-level, ontologies are designed to consist of highly general terms and relational expressions that are common across all domains. Upper ontologies play a critical role in enabling interoperability between heterogeneous data by providing a common backbone for other domain-specific ontologies \citep{mascardi2007comparison}. By extending an upper ontology to develop domain-specific ontologies, they can share common root terms, which allow the data curated to be shared via abstraction to those general terms.

An upper ontology also provides a framework to categorize and constrain different domain-specific entities, e.g., distinguishing between independent entities and dependent entities. Also, it provides existing theories that could be extended to domain-specific entities and ensure that the data modeled with them follow the logical consistencies imposed by the upper ontology. Finally, an upper ontology serves as the natural starting point for defining new terms for domain-specific entities.

DOLCE, or Descriptive Ontology for Linguistic and Cognitive Engineering, is an upper ontology with ``a clear cognitive bias" \citep{gangemi2002sweetening}. That is, it ``aims at capturing the ontological categories underlying natural language and human commonsense". While the founder of DOLCE has indicated that a formal ontology should reflect reality \citep{guarino1998formal}, this description suggests that DOLCE is an ontology of \emph{concepts} formulated by humans to interpret reality. DOLCE has a rich set of axiomatizations and found several successful applications in knowledge-based systems \citep{mascardi2007comparison}.

BFO, or Basic Formal Ontology \citep{smith2015bfo}, is an upper ontology of universals with strict commitment toward ontological realism, as stated earlier. It started as an ontology to represent ``dynamic features of reality" \citep{grenon2004snap} and found significant success in the biomedical domain \citep{grenon2004biodynamic}, e.g., in the development of Gene Ontology \citep{ashburner2000gene} and the establishment of Open Biomedical Ontologies (OBO) Foundry \citep{smith2007obo}. It shares a number of similar categorizations as DOLCE, except a few notable differences such as in material constitution and quality descriptions.

GFO, or General Formal Ontology \citep{herre2010general}, is an upper ontology that is explicitly stated as an ontology for ``conceptual modeling" and features a rich set of axiomizations, similar to DOLCE. The focus of applications has been in the biomedical domain alike BFO. 

YAMATO, or Yet Another More Advanced Top-level Ontology \citep{mizoguchi2010yamato}, is a foundational ontology created to address some of the issues identified by its author from existing upper ontologies. Compared to the above three ontologies, YAMATO defines much more in-depth categories intended to be more useful in applications. However, our experience has shown such detailed categories can conflict with the partitioning of the domain-specific categories required for a particular application, making the adoption of the ontology difficult. In the end, one may have to disregard much of the detailed categories developed for the upper ontology.

SUMO, or Suggested Upper Merged Ontologies \citep{niles2001towards}, is essentially an aggregate of various upper level ontologies, including the first three ontologies mentioned above. It is the largest public ``upper" ontology in terms of its contents \citep{mascardi2007comparison}. Similar to YAMATO, such extensive contents actually make it difficult to be used for developing domain-specific ontologies.

Among the five upper ontologies reviewed, only BFO is explicitly stated as a realist ontology. Hence, although other upper ontologies could be capable of adequately representing physical phenomena for simulation, we have hypothesized that the theories developed for BFO are more likely to be relevant because of its realist principles. In addition, BFO has demonstrated success in the scientific domain and in particular, representing physics in biology \citep{cook2008bridging,cook2011physical,cook2013ontology}. While DOLCE has been successfully extended to represent various manufacturing and design related concepts \citep{borgo2007foundations,borgo2009artefacts,borgo2009formal,sanfilippo2015feature,sanfilippo2015towards}, it should be emphasized that the nature of such concepts can be quite different from entities involved in physics -- the former are mostly social constructs while the latter are more of \emph{brute facts} \citep{searle1995construction}, the type of entities that have been the primary focus of BFO in the past. Lastly, but perhaps most importantly, there are well-documented resources on how to use BFO to develop domain-specific ontologies \citep{arp2015building, smith2015bfo}, as well as an active discussion group\footnote{https://groups.google.com/forum/\#!forum/bfo-discuss} where questions about BFO can be discussed. For these reasons, BFO was chosen as the upper ontology to develop PSO.

%--------------------------------------------------

\subsection{Ontology applications for CAD / CAE / PLM}

Several applications of ontologies for engineering design can be found, particularly to solve varying challenges for CAD, CAE, or PLM software. 

Ontologies and formal data models have been developed as attempts to capture additional semantics beyond geometries in CAD software. \cite{horvath1998towards} used an ontological approach to extend the concept of ``features'' in engineering design to not only represent forms or shapes, but to also capture other domain knowledge considered in design. This approach has been followed by several endeavors in creating more rich and formalized data models to capture design semantics, e.g., Core Product Model \citep{fenves2008cpm2} and OntoSTEP \citep{barbau2012ontostep} developed by National Institute of Standards and Technology. A number of efforts involved applying mereotopological theories to formalize product assembly information, e.g., \cite{kim2006ontology}, \cite{kim2008mereotopological}, \cite{demoly2012mereotopological}, and \cite{gruhier2016theoretical}. Another group of work has used DOLCE to axiomatize various notions used in engineering design such as artifacts \citep{borgo2009artefacts}, features \citep{sanfilippo2015feature}, products \citep{sanfilippo2015towards}, and manufacturing-related entities \citep{borgo2007foundations}.

Ontologies have also been applied to improve interoperability between applications, mainly within the PLM context. \cite{young2007manufacturing} applied logically rigorous ontologies such as PSL \citep{gruninger2003process} to formalize the meanings of the terms used in PLM software so that sharing of manufacturing knowledge can be maximized. Open Assembly Model \citep{fiorentini2007towards} is an extension of the Core Product Model to support the exchange of assembly and tolerance information among PLM applications. \cite{matsokis2010ontology} developed an ontology based on OWL to solve data integration and interoperability challenges in closed-loop PLM.

Ontologies have also been used in various CAE contexts. \cite{grosse2005ontologies} developed an ontology to categorize engineering analysis models to support their reuse and sharing. \cite{witherell2007ontologies} developed an ontology to capture knowledge related to engineering design optimization for its sharing and reuse, e.g., the modeler's rationale and justification behind the choice of optimization models. In addition, \cite{freitas2014towards} proposed an ontology-driven web-based platform for sharing finite element method (FEM)-based simulation models. 

Besides for capturing the knowledge used in simulation or optimization models, \cite{sun2009framework} demonstrated the application of an ontology to automate FEM problem definition and analysis. The work done by \cite{benjamin2006using} and \cite{turnitsa2010ontology} have also proposed using ontologies to assist in modeling simulation problems, although no ontology was actually presented in their work. Also, \cite{gruhier2015formal} developed an ontology to support assembly sequence planning and \cite{arena2017methodological} used an ontology to instantiate Petri-net models for manufacturing process simulation. 

In line with most of the prior work, PSO aims to solve interoperability challenges among physics-based simulation applications. Similar to the work done by Borgo and his colleagues, we use an upper ontology, in our case BFO, as the basis for defining the terms of PSO. In addition to improving interoperability, we intend PSO to help users set up physics-based simulation problems. This intention is much like the work of \cite{sun2009framework}, with the difference being that their ontology captures FEM constructs while our ontology prioritizes capturing the actual physical phenomena of interest. Finally, an important distinction can be made between the work done by \cite{grosse2005ontologies} and our work. The former work developed an ontology to categorize different types of simulation models. Our work focuses on developing ontologies to represent the physical phenomena to be simulated.

%-------------------------------------------------

\subsection{Other work related to physics and simulation}

In the artificial intelligence community, the idea of \emph{na\"ive physics} have been explored to replicate the common sense reasoning of humans \citep{hayes1978naive}. The representation methods chosen for naive physics tended to simplify the details required in classical physics while focusing more on the formalism required for efficient reasoning \citep{de1984qualitative}. On the other hand, work such as \cite{randell1992naive} has illustrated the importance of semantics in the required formalism.

\cite{borst1997engineering} developed an ontology called PHYSSYS to encompass the domains of systems modeling, simulation, and design. It incorporated theories from mereotopology and systems engineering to create three views of physical systems -- component, process, and engineering mathematics. As described later in the current paper, the overall structure of their ontology is quite similar to PSO, which inherited the main branches from BFO -- independent continuants, processes, and information content entities. In addition, PSO includes the branch of specifically dependent continuants (e.g., qualities) from BFO. On the other hand, the PHYSSYS ontology aims to model and simulate physical systems using ordinary differential equations, which is a simplified view of physics compared to using partial differential equations as in the current work. In addition, the ontology developed nor its detailed documentation could not be found anywhere for use.

Specific to the domain of physics, Collins \citep{collins2004standardizing,collins2014towards} has indicated working towards developing a general ontology for physics. However, we are not aware of any ontology developed. In the biomedical domain, Cook and his colleagues have been building an ontology for describing physical properties, processes, and dependencies in biology \citep{cook2008bridging,cook2011physical,cook2013ontology}. Their work was extended from BFO, alike ours, and GFO.

%=================================================

\section{Scope and requirements of PSO}
\label{scope}

The goal of PSO is to identify some aspects of reality through the common perspective held in physics-based simulation. Consequently, the physical phenomenon modeled with PSO can be translated into information that can be shared across different solvers and help set up valid simulation problems for specific solvers.

First, the perspective assumed is based on classical mechanics. Therefore, PSO needs to be able to identify three-dimensional objects at the macroscopic level moving at speeds much slower than the speed of light, and the relevant properties of objects at a particular time point (such as shape, mass, velocity, and energy). PSO also needs to account for categorizing processes in which the properties of objects change over time, while distinguishing different types of physical behaviors occurring simultaneously, e.g., structural behavior vs. thermal behavior of an object. In doing so, PSO reflects the laws of physics assumed in classical mechanics. Note that PSO does not have to reason about changes that occur during physical behaviors, because such changes are exactly what simulation solvers are designed to compute. PSO simply needs to capture the snapshots of physical phenomena that can be used to provide the sufficient information for initiating simulation solvers. Finally, PSO is not intended to support the views of quantum mechanics, quantum field theory, or relativistic mechanics.

In addition, the current work must consider the fact that most physics-based simulation techniques entail modeling physical behaviors as boundary value problems, consisting of partial differential equations (PDE) and a set of constraints. Typically, a PDE describes a particular law of physics that governs the behavior of an object. Once one or more physical behaviors of interest are identified, setting up and solving a simulation problem involves specifying the domain(s) of equations, boundary (and initial) conditions, and certain parameters of equations. Hence, the competency of PSO can be evaluated by whether these problem specifications can be derived from the description of physical phenomena modeled with PSO.

\subsection{Heat transfer example}
\label{heat}

Modeling the heat transfer behavior of an object as a PDE-based boundary value problem is presented below, and illustrated in Figure~\ref{fig2}. Then, the problem specifications required for solving this type of boundary value problems are generalized.

\begin{figure}[b]
\begin{center}
\includegraphics[width=0.3\linewidth]{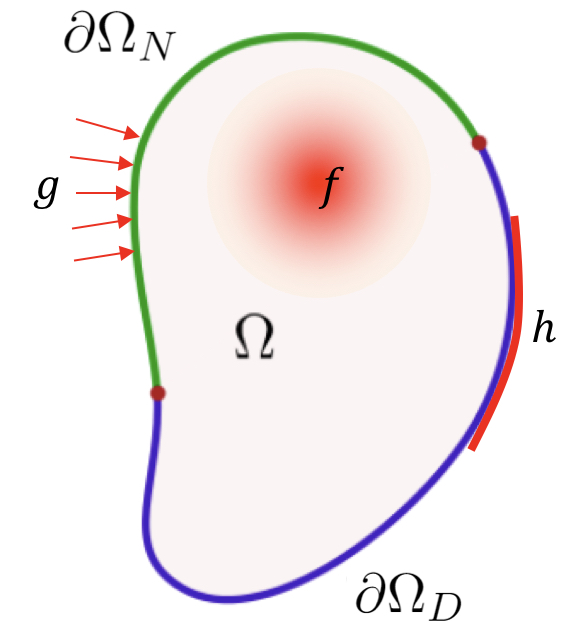}
\caption{Heat transfer problem example involving Dirichlet and Neumann boundary conditions.}
\label{fig2}
\end{center}
\end{figure}

A heat equation describing the distribution of heat in a given body, represented by the domain $\Omega$, over time $t$, can be stated as follows:
\begin{equation} 
\rho c_p \dfrac{\partial u}{\partial t} - \nabla \cdot (\kappa \nabla u) = f \text{, in } \Omega 
\end{equation}

Here, $u$ is a function $u : \Omega \times [0,T] \rightarrow \mathbb{R}$ that represents the temperature of the body over a time period $T$. The parameters $\rho$, $c_p$, and $\kappa$ are material properties of the body (density, specific heat capacity, and thermal conductivity, respectively). $f$ is known as a source term that defines a volumetric heat source over the body. 

Constraints are imposed on the solutions of the equation, as boundary conditions:
\begin{gather} 
u = h \text{ on } \partial\Omega_D \\
\kappa \dfrac{du}{dn} = g \text{ on } \partial\Omega_N \, .
\end{gather}

The two main types are known as Dirichlet and Neumann boundary conditions, in the order presented above. These conditions are applied on the subsets of the boundary surface of the body, denoted as $\partial\Omega_D$ and $\partial\Omega_N$, respectively. The Dirichlet boundary condition specifies that the temperature at $\partial\Omega_D$ is equal to some prescribed temperature, $h$. The Neumann boundary condition specifies that $du/dn$, the spatial rate of change of temperature in the normal direction to $\partial\Omega_N$, is proportional to some prescribed temperature flux, $g$. This models conductive heat transfer across the surface. One could also have a Robin boundary condition, which is a combination of Dirichlet and Neumann boundary conditions, that can be used to describe temperature-dependent surface fluxes for example.  

We also have initial conditions such as:
\begin{equation} 
u = u_\circ \text{ at } t=0 \text{, in } \Omega
\end{equation}
This condition specifies the initial temperature of the body to be \(u_\circ\).

Simulating a heat transfer behavior entails solving the equations stated above for the unknown values of $u$ throughout the body $\Omega$ over a period of time, tracked by $t$. In addition, in order to have a well-posed problem, the following conditions must be met. First, $\partial \Omega_D \cup \partial \Omega_N = \partial \Omega$ must be true, which means the entire boundary of the body must be prescribed with one of the two boundary conditions. Also, $\partial \Omega_D \cap \partial \Omega_N = \emptyset$ must be true, which means that no surface on the body can have more than one boundary condition. Finally, there must be an initial condition specified to find a unique solution to the problem. These conditions reflect the laws of physics for heat transfer and establish the requirements for what needs to be entailed by the ontology. 

The association between the elements of the equations and their corresponding physics entities in reality can be summarized as follows:
\begin{itemize}
\item Domain, $\Omega$: Physical object(s) of interest
\item Boundaries, $\partial\Omega_D, \partial\Omega_N$, and $\partial\Omega_R$: Surfaces of the object
\item Time, $t$: Time associated with the physical phenomenon
\item Material parameters, $\rho, c_p,$ and $\kappa$: Material properties of the object
\item Physical parameters, $u, u_\circ, f, g,$ and $h$: Physical properties of the object and its surfaces
%\item $u$: Temperature inside the domain
%\item $u_\circ$: Initial temperature inside the domain
%\item $f$: Volumetric heat source in the object
%\item $g$: Heat flux across the boundary $\partial \Omega_N$
%\item $h$: Temperature of the boundary $\partial\Omega_D$
\end{itemize}

\subsection{Other physics and multi-physics consideration}

Other physical behaviors, such as structural, fluid, or electromagnetic behavior, can be described using different PDEs. Yet, the elements of those different equations will correspond to similar types of physics entities in reality as in the heat transfer example. The domain will typically refer to physical objects and the boundaries will refer to the surfaces of those objects. Depending on the physical behavior of interest, different types of material properties and physical properties will be relevant. For example, the following is the governing equation for a structural behavior (linear elasticity): 
\begin{gather} 
\nabla\cdot\sigma(u) = f \text{ in } \Omega \\
\sigma(u) = A(\lambda, \mu) : \varepsilon(u)
%\sigma = \lambda\,\text{tr}\,(\varepsilon(u)) I + 2\mu\varepsilon(u) 
\end{gather}
For this equation, $u$ and $f$ represent the physical properties of displacements and a body force, respectively, and $\lambda$ and $\mu$ are material properties related to linear elasticity.

A problem modeled could involve multiple physical behaviors occurring at the same time and interacting with each other. For instance, the thermal behavior of an object can affect its structural behavior, because the temperature differences in the object can cause deformations of the object. For example, the following equations demonstrate the interaction between thermal and structural behaviors: 
\begin{gather} 
\rho c_p \dfrac{\partial u_1}{\partial t} - \nabla \cdot (\kappa \nabla u_1) = f_1 \text{, in } \Omega
\end{gather}
\begin{gather}
\nabla\cdot\sigma(u_1, u_2) = f_2 - \alpha \nabla u_1 \text{ in } \Omega \\
\sigma(u_1, u_2) = A(\lambda, \mu) : \varepsilon(u_2) - \alpha (u_1-u_1^\text{ref}) I
%\sigma(u_1, u_2) = \left(\lambda\,\text{tr}\,(\varepsilon(u_1)) - \alpha (u_2-u_2^\text{ref})\right) I + 2\mu\varepsilon(u_1)
\end{gather}
Here, the physical properties $u_1$ and $u_2$ represent temperature and displacements, and $f_1$ and $f_2$ represent a volumetric heat source and a body force, respectively. In this scenario, the linear static equation (8-9) now depends on the solution of the thermal equation (7), namely $u_1$. Hence, the two equations are coupled, and such dependencies between physical properties should be identified in a problem model.

%\abcomment{Would it be worthwhile to introduce a second example, namely the equations for a coupled Thermal-Structural problem? I could write those down for you.}

%\abcomment{There are lots of ways in which a multiphysics problem can be coupled.  Are all of them captured here?  Maybe we should go through the Thermal/Structural problem as a test case.  But even in single-physics, you can impose constraints as well as boundary conditions e.g. in a structural problem you can impose that the $x$-component of the displacement is fixed (Dirichlet) while the $y$- and $z$-components are free (Neumann).  You can also couple a deformable body to a beam at a point.  Technically, I would characterize this as a multiphysics problem since we would be solving the linear elastic PDE coupled to the beam ODE.  But still, I wonder if this sort of thing can be captured in the ontology so far.}

\subsection{Discretization required for simulation}

The key step in using a computational algorithm to approximately solve physics problems, formulated as in the above examples, is discretization. The PDEs that represent physical behaviors in their original forms involve continuous variables. However, in order to solve for those variables, a solver must discretize the equations in both space and time, and turn the problem into iterations of linear algebra problems of the form $A_n x_n = b_n$. (For more information on this topic, see \cite{langtangen2016solving}). Examples of discretization include the use of a mesh to discretize the domain in a finite element method or the integration of equations over a small time step, known as temporal discretization.   

\subsection{Capturing the physics and its simulation with an ontology}

Now we discuss how an ontology can capture the relevant physics and provide the required input to physics-based simulation. First, an ontology should carve out the portions of reality that are relevant to the descriptions of physics presented in Sections 3.1-3.2. Based on our example, such partitioning should identify different physical behaviors occurring, the objects involved in those behaviors, the boundaries of those objects, the physical properties of the objects and boundaries, and so on. Once those physics entities are identified, an ontology should interpret them in the forms that can be understood by simulation solvers. These interpretations involve further partitioning, e.g., discretization of the representation of an object being simulated, where such interpretations are solver-specific.

In summary, the first partitioning step focuses on capturing the physical phenomena observed through the view of classical mechanics and PDEs, while the second partitioning step, which involves solver-specific interpretations, refers to the model captured in the first step. Hence, PSO is developed in two parts -- the terms of PSO-Physics, which are used to model the physical phenomenon, and the terms of PSO-Sim, which represent the information used to interpret the physical phenomenon modeled.

%=================================================

\section{PSO-Physics}

\begin{figure}[H]
\begin{center}
\includegraphics[width=\linewidth]{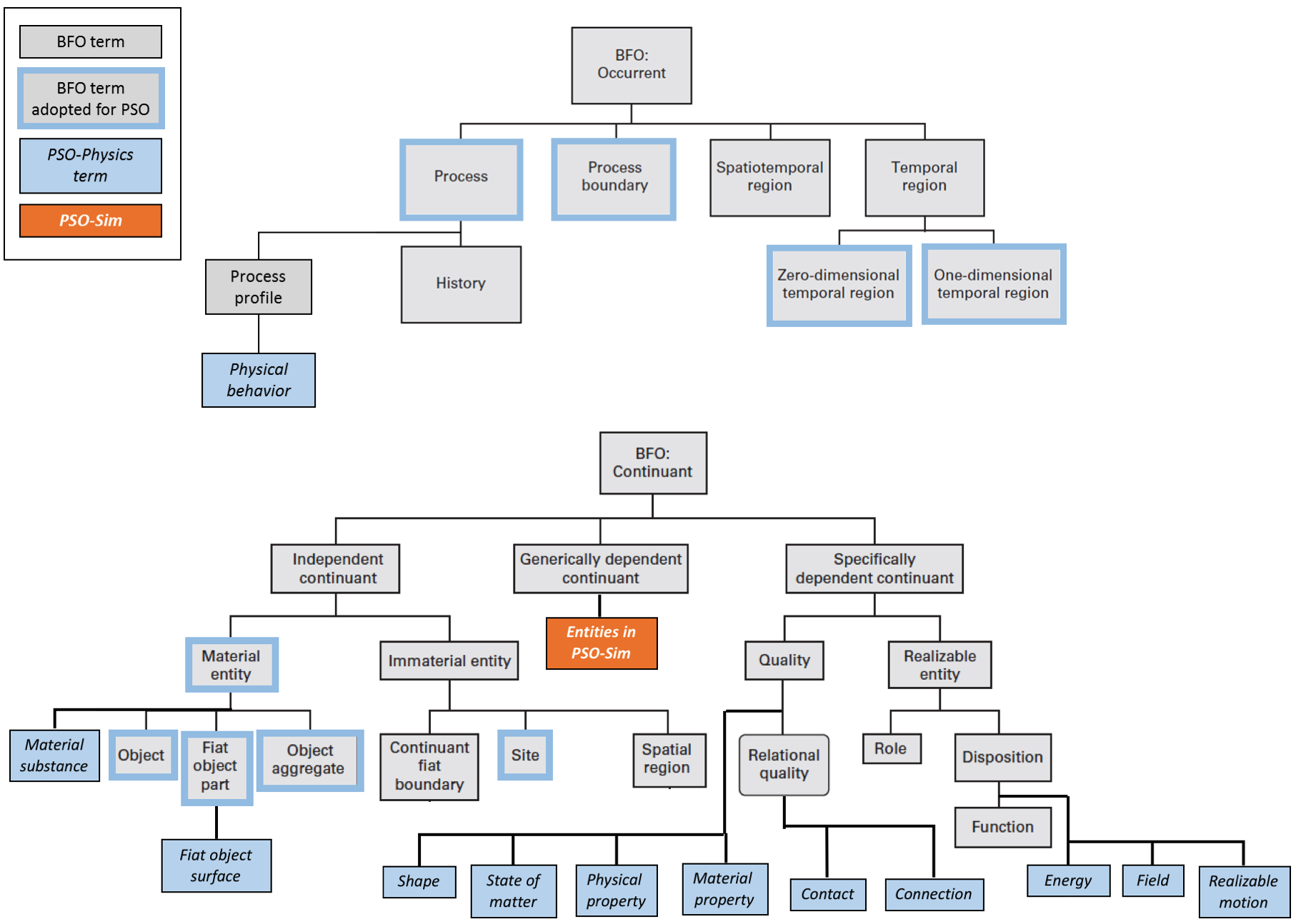}
\caption{PSO terms categorized in the BFO hierarchy. BFO terms adopted for PSO have borders highlighted. Original PSO terms are in italics.}
\label{fig3}
\end{center}
\end{figure}

The current section presents the terms of PSO that capture the entities and relations found in physical phenomena, which are relevant in formulating the required information for physics-based simulation. These terms are denoted as PSO-Physics. Many of the terms are directly adopted from BFO and those terms are denoted with a prefix, {\small BFO}. Terms that are original to the current work, but categorized under BFO terms, are denoted with a prefix, {\small PSO}. In the current section, only the terms that are directly relevant to the motivating example in Section 3 and the case study problem in Section 4 are presented. Other terms of PSO-Physics that could be potentially useful in the future are presented in Appendix \ref{appx:A}.  Also, natural language definitions are provided in the current work while formal axioms will be completed in future work. Figure~\ref{fig3} shows the categorization of the PSO-Physics terms under the BFO hierarchy.

%-------------------------------------------------

\subsection{Entities in PSO-Physics}

Entities are presented in three groups based on the categorization used in BFO. The first group, named \emph{occurrents}, consists of entities related to processes, e.g., the thermal behavior of an object or the temporal interval of that thermal behavior. The second group consists of  \emph{independent continuants}, which endure their identities while bearing certain qualities that change over time, e.g., the object of interest and its parts. The last group denotes \emph{specifically dependent continuants}, which all depend on one or more independent continuants for their existence, e.g., the temperature of the object or the thermal conductivity of its material substance.

\subsubsection{Occurrents}

In BFO, an occurrent is ``an entity that unfolds itself in time, or it is the instantaneous boundary of such an entity [...] along the time axis, or it is a temporal or spatiotemporal region that such an entity occupies'' \citep{arp2015building}. The following occurrent entities, most of them adopted from BFO, are used in PSO-Physics. 

\paragraph*{{\small BFO: }\textit{Process.}} 
A process is ``an occurrent entity that exists in time by occurring or happening, has temporal parts, and always depends on some (at least one) material entity" \citep{arp2015building}. In PSO, a process can be used to demarcate the physical process involving one or more objects to be simulated, e.g., the pumping process of a pump assembly. A physical process can be further demarcated into parts based on the different \emph{physical behaviors} associated with the process, as defined below.

\paragraph*{{\small PSO: }\textit{Physical behavior.}}
A physical behavior demarcates some parts of a process according to specific laws of physics. For example, the thermal behavior of fluid, the dynamic motion behavior of a turbine, and the electromagnetic behavior of an electric motor could be identified as different physical behaviors of a pumping process.

\cite{smith2012process} denotes such parts of a process as \emph{process profiles}. Process profiles can be thought as different aspects of a process that are identified for a particular purpose. The notion of \emph{process profile part} is different from \emph{temporal part}, as in a process profile occurs throughout the same temporal region as the whole process. Depending on the types of simulation in which a user is interested, different physical behaviors of a physical process can be selectively identified. The following is a definition of a physical behavior:\\

\begin{minipage}{0.925\textwidth}
{\small PSO: }\textit{physical behavior} = def. a {\small BFO: }\textit{process profile} that is part of some process while occupying the same temporal region, and follows a specific law of physics.
\end{minipage}

\paragraph*{{\small BFO: }\textit{1-D temporal region.}}

A one-dimensional temporal region is equivalent to a temporal interval \citep{arp2015building}, which can be used to identify the duration of a physical process, e.g., the duration of a pumping process.

\subsubsection{Independent continuants}

An independent continuant is an entity ``that continues to exist through time'' and ``is the bearer of qualities'' \citep{arp2015building}. One main type of an independent continuant in BFO is a \emph{material entity}, which is defined to have ``some portion of matter as part'' \citep{arp2015building}. 

\paragraph*{{\small PSO: }\textit{Material substance.}}

Under a material entity, we introduce a category called \emph{material substance} to classify different types of material substances such as aluminum, concrete, or ethanol, which are used to identify what the object to be simulated is made of. In engineering, we typically denote material substances simply as ``materials''. In case of a multi-material object, multiple material substances can be associated with the particular object. A material substance is defined as follows: \\

\begin{minipage}{0.925\textwidth}
{\small PSO: }\textit{material substance} = def. a chemical substance that a {\small PSO: }\textit{material entity} is made of.  \\
\end{minipage}

The definition of the \textbf{made of} relation is provided in Section 4.2.2. We use a parthood relation to associate a material substance to a material entity. This approach was inspired from DOLCE \citep{guarino2000identity, borgo2009artefacts}, which assumes the material constitution theory \citep{rea1995problem}. \\

Other categories under a material entity include the three entities defined in BFO, \emph{object}, \emph{object aggregate}, and \emph{fiat object part}. These terms are reused in PSO with the same definitions as BFO. In addition, we define a term called \emph{fiat object surface} to denote a special type of fiat object part.

\paragraph*{{\small BFO: }\textit{Object.}}

In BFO, an object is a material entity that is ``spatially extended in three dimensions [and] causally unified'' \citep{arp2015building}. It can be used to identify physical objects being simulated, such as a turbine, a beam, a portion of exhaust gas, or a portion of blood.

\paragraph*{{\small BFO: }\textit{Object aggregate.}}

An object aggregate is a ``material entity that is made up of a collection of objects and whose parts are exactly exhausted by the objects that form this collection'' \citep{arp2015building}. It can be used to identify a group of physical objects to be simulated together, such as a portion of oil flowing through a pipe, a four-bar mechanism consisting of three linkage members and the ground, or an electric rotor rotating around a stator. A mechanical assembly, which is an entity of special interest in multi-body dynamics, can be thought as an object aggregate with constrained relative motions between each pair of objects.

\paragraph*{{\small BFO: }\textit{Fiat object part.}}

A fiat object part is a ``material entity that is a proper part of some larger object, but is not demarcated from the remainder of this object by any physical discontinuities'' \citep{arp2015building}. It can be used to identify some portion of a physical object that might be of special interest, such as a portion of fluid in turbulence, a part of a beam under large deformations, or a part of a metal rod under the influence of a electromagnetic field. In addition, a fiat object part can be used to identify different regions of a multi-material object that are made up of different material substances, such as the layers in a laminated composite object.

\paragraph*{{\small PSO: }\textit{Fiat object surface.}}

A fiat object part is extended to define an important category in PSO. A fiat object surface is defined as follows: \\

\begin{minipage}{0.925\textwidth}
{\small PSO: }\textit{fiat object surface} = def. a {\small BFO: }\textit{fiat object part} of a {\small BFO: }\textit{object} that is minimal in one spatial dimension. \\
\end{minipage}

Object surfaces are especially important in PSO because they are referred by boundary conditions to set up a valid problem for simulation and identify contacts between objects. Because a fiat object surface is a fiat object part (hence a material entity), it can bear physical and material properties. It is different from a \emph{continuant fiat boundary} in BFO, which is an immaterial entity \citep{arp2015building}. Rather, one could define a fiat object surface as a portion of an object that is located along a particular continuant fiat boundary. What constitutes as \emph{minimal} can be defined based on the modeler's perspective on a particular physical phenomenon of interest. The important point here is that once a fiat object surface is demarcated based on that perspective, it can be communicated between different applications in a consistent manner. Examples of a fiat object surface include the layer of water on top of a lake, the interior wall of a pipe, or the outer surface of a tire in contact with the ground.

\paragraph*{{\small BFO: }\textit{Site.}}

A site is different from the above types of independent continuants in that it is not made of any material substance. In BFO, a site is a type of an \emph{immaterial entity} and defined as ``a three-dimensional immaterial entity that either (1) is (partially or wholly) bounded by a material entity or (2) is a three-dimensional immaterial part of an entity satisfying (1)'' \citep{arp2015building}. A prototypical example of a site is a hole, and sites usually contain other material entities. In PSO, a site can be used to identify a hole on a metal block through which a piston translates, a channel through which a portion of water flows, or a vacuum.

\subsubsection{Specifically dependent continuants}

\noindent Next, two types of specifically dependent continuants are presented. These are continuants that depend on one or more other independent continuants for their existence. The first type is a \emph{quality}, which inheres in an independent continuant bearer and is fully exhibited or realized. The second type is a \emph{relational quality}, which is a quality that has multiple independent continuants as its bearers. A \emph{realizable entity}, which inheres in an independent continuant bearer but is exhibited or realized only through certain processes, is another type found in BFO and its extensions in PSO are presented in Appendix A. These definitions and categorizations directly come from BFO \citep{arp2015building}.

\subsubsection*{\underline{Qualities}}

\paragraph*{{\small PSO: }\textit{Physical property.}}

Physical properties are a set of qualities that describe the physical state of a material entity. \\

\begin{minipage}{0.925\textwidth}
{\small PSO: }\textit{physical property} = def. a {\small BFO: }\textit{quality} that determines the physical state of a {\small PSO: }\textit{material entity}, and can be measured as quantitative values based on some measurement units. \\
%\begin{flushleft} $ \forall x \ PhysicalProperty(x) \to \exists y \ MaterialEntity(y) \land s\_depends\_on(x, y) $ \end{flushleft}
\end{minipage}

Examples include the temperature of a heat sink, the velocity of a moving ball, or the pressure applied on the wall of some fluid. These physical properties are typically what get computed during physics-based simulation to predict the behavior of objects. In the current work, we have not made distinctions of physical properties into those that are vector vs. scalar quantities, or fundamental vs. derived quantities. Instead, we allow such distinctions to be made for specific applications of PSO. The current work also does not present an exhaust list of physical properties, but existing ontologies of physical quantities such as \cite{haasquantities} and \cite{lefort2005ontology} can be imported to populate the category.

\paragraph*{{\small PSO: }\textit{Material property.}}

Material properties characterize a material substance, which in turn affect how a material entity made of the material substance physically behaves under some physical processes. \\

\begin{minipage}{0.925\textwidth}
{\small PSO: }\textit{material property} = def. a {\small BFO: }\textit{quality} that can be measured to identify the physical characteristics of a {\small PSO: }\textit{material substance}. \\
% \begin{flushleft} $ \forall x \ MaterialProperty(x) \to \exists y \ MaterialSubstance(y) \land s\_depends\_on(x, y) $ \end{flushleft}
\end{minipage}

Examples of material properties include density, viscosity, elastic modulus, thermal conductivity, opacity, etc. Material properties are distinguished from physical properties because the values of the former properties can be used as the identity criteria for a material substance while the latter cannot. For example, the value of an elastic modulus alone can dictate the type of material substance that a material entity is made of. However, the pressure applied on a material entity cannot dictate which material substance it is made of.

Again, we do not present a detailed taxonomy of material properties in the current work, but assume that a specific application of PSO can define one for its purpose.

\subsubsection*{\underline{Relational qualities}}

\paragraph*{{\small PSO: }\textit{Contact.}}

A contact is formed between a pair of surfaces of objects, where the two surfaces are touching with each other but there is no overlap between the objects. In mereotopology, this relationship is expressed as an \emph{external connection} \citep{varzi1996parts,cohn1997qualitative,smith2000fiat}. \\

\begin{minipage}{0.925\textwidth}
{\small PSO: }\textit{contact} = def. a {\small BFO: }\textit{relational quality} that is formed between a pair of {\small PSO: }\textit{fiat object surfaces}, where they are externally connected to each other. \\
\end{minipage}

Examples of a contact include teeth of gears engaged, a water droplet sitting on a glass panel, or a protective membrane glued on a device. The last example can be thought as a \emph{strong} rather than a \emph{weak} contact \citep{smith1996mereotopology}, where the objects in contact will move together unless a certain separating process occurs.

%\paragraph*{{\small PSO: }\textit{Connection.}}

%A connection is formed between a pair of overlapping objects where the objects share a common object fiat surface. It is equivalent to the \emph{connection} described in mereotopology \citep{smith1996mereotopology,varzi1996parts}. \\

%\begin{minipage}{0.925\textwidth}
%{\small PSO: }\textit{connection} = def. a {\small BFO: }\textit{relational quality} that is formed between a pair of overlapping {\small BFO: }\textit{objects}, where the objects share a common {\small PSO: }\textit{fiat object surface}. \\
%\end{minipage}

%An example of a connection is that created between two welded pipes. The fixed connections typically referred in engineering created with bolts, screws, or adhesives, etc., are not considered as a {\small PSO: }\textit{connection}, because no surfaces are shared between objects in such cases. Rather, they can be treated as in a strong contact, and the objects involved have \emph{realizable motions} that are dependent on each other.

\subsection{Relations in PSO-Physics}

PSO-Physics includes a small set of primitive relations to link different types of instances during modeling. Most of these relations are adopted from BFO. 

\subsubsection{Relations from BFO}

The following is a list of relations from BFO that is adopted in PSO-Physics. The details and logical definitions of these relations can be found in \cite{smith2015bfo}.

\paragraph*{\textit{occupies temporal region.}}

This relation is held between an instance of a occurrent and an instance of a temporal region. For example, a pumping process \textbf{occupies temporal region} of a 5-second time interval.

%\paragraph*{\textit{temporal part of.}}

%This relation is held between two instances of occurrents, where the former is a sub-phase of the latter. For example, an ignition process is \textbf{temporal part of} an engine starting process.

\paragraph*{\textit{process profile of.}}

This relation is held between an instance of a process profile and an instance of a process, where the former is a particular, identifiable aspect of the latter. For example, a thermal behavior of an engine during engine ignition is a \textbf{process profile of} an engine ignition process.

\paragraph*{\textit{has participant.}}

This relation is held between an instance of a process and an instance of the continuant involved in the process. For example, an oil pumping process \textbf{has participant} a pump. 

\paragraph*{\textit{continuant part of.}}

This relation is held between two instances of continuants to describe parthood between them. For example, a swingarm is \textbf{continuant part of} a motorcycle.

\paragraph*{\textit{located in.}}

This relation is held between two instances of continuants to describe their relative locations. For example, a portion of fluid is \textbf{located in} a fluid channel.

\paragraph*{\textit{s-depends on.}}

This relation is held between an instance of a specifically dependent continuant and an instance of the independent continuant bearer. For example, the temperature of a pipe \textbf{s-depends on} the pipe.

\subsubsection{New relations for PSO-Physics}

Only two new relations are introduced in PSO-Physics. The first relation is the result of our commitment to the material constitution theory, and the second relation is needed to express dependencies between specifically dependent continuants due to physical laws or other physical constraints in reality.  

\paragraph*{\textit{made of.}} 

This relation is held between an instance of a {\small PSO:} material entity and an instance of a {\small PSO:} material substance. For example, an engine block is \textbf{made of} a portion of cast iron.
% \begin{equation*}
%    \forall x,y \ made\_of(x, y) \to MaterialEntity(x) \land MaterialSubstance(y) \land part\_of(y, x)
% \end{equation*}

\paragraph*{\textit{physically related to.}}

This relation is held between two instances of specifically dependent continuants, to express those two instances are related to each other due to certain laws of physics. For example, the temperature of the air in this room is \textbf{physically related to} the pressure of the air in this room. The relation is transitive and symmetric.
%\begin{equation*}
%     \forall x,y \ physically\_related\_to(x, y) \to SDC(x) \land SDC(y) \land (x \neq y)
%\end{equation*}
% where SDC is a specifically dependent continuant. 

%=================================================

\section{PSO-Sim}

So far, all the terms in PSO-Physics correspond to certain physics entities in reality. With PSO-Physics, it is hypothesized that a user can adequately model the particular physical behavior of interest. To interpret this model as input to simulation solvers, we need to introduce information content entities (ICEs) \citep{ceusters2015aboutness}. ICEs, in BFO categorization, are generically dependent continuants that are about something. In our case, ICEs are about the physics entities identified with PSO-Physics. Few examples of ICE terms in PSO-Sim relevant to typical simulation solvers are presented in the current section. While most of these terms have specific meanings in the mathematical context, here we provide some elucidations of how they are related to the physical phenomena modeled with PSO-Physics.

%-------------------------------------------------

\subsection{Entities in PSO-Sim}

%\paragraph*{{\small PSO: }\textit{Simulation problem definition.}}

%A simulation problem definition (SPD) is an aggregate of ICEs that are required for a particular physics-based simulation solver. An instance of a SPD contains, for example, contains instances of the ICEs presented below. A SPD is about the physical behavior(s) to be simulated.

\paragraph*{{\small PSO: }\textit{Domain.}}

A domain represents some independent continuant that participates in the physical behavior to be simulated. For example, a domain can represent a solid object in structural simulation, a fluid substance in fluid simulation, or a spatial region in which a field is located in electromagnetic simulation.

\paragraph*{{\small PSO: }\textit{Geometric model.}}

A geometric model represents the shape of some independent continuant. In CAD / CAE applications, geometric models are typically concretized in geometry files such as STEP or OBJ files. Note the distinction of geometric models from the physical entities themselves -- the former is an information content entity that is a visual representation of the latter.

\paragraph*{{\small PSO: }\textit{Mesh.}}

A mesh is a specific type of a geometric model that uses a set of polygons to represent the shape of some independent continuant. It is one of many techniques used to discretize the space of a domain. 

\paragraph*{{\small PSO: }\textit{Boundary condition.}}

A boundary condition refers to a \emph{situation} in which some fiat object surface of a material entity bears some physical property throughout the physical behavior of interest. As presented in Section \ref{scope}, the physical properties vary depending on the types of boundary conditions and physical behavior. In simulation, boundary conditions are always applied on the domain boundaries, which represent the surfaces of the corresponding independent continuant in physical reality.

The notion of a situation is adopted from GFO. GFO defines a \emph{configuration} as an aggregate of facts formed with multiple material structures, properties, and relations \citep{herre2005ontology}. Then, a situation is defined as ``a special configuration which can be comprehended as a whole and satisfies certain conditions of unity, which are imposed by relations and categories associated with the situation'' \citep{herre2005ontology}. BFO does not explicitly categorize a situation, but \cite{ceusters2015aboutness} uses ``configuration", defined as a portion of reality made up by multiple entities that are related to each other. Because it was not clear how to categorize a situation under BFO, it has not been included as part of PSO. 

Another issue is related to the limited expressiveness of first-order logic. A proper logical definition of a boundary condition would require reification of relations involved in a situation, as shown in Figure 4(a). However, quantifying over relations is not possible in first-order logic. Therefore, we are limited to a weaker definition of a boundary condition shown in Figure 4(b), which does not involve reification of relations. 

\begin{figure}[t]
\begin{center}
\subfigure[]{\includegraphics[width=0.58\linewidth]{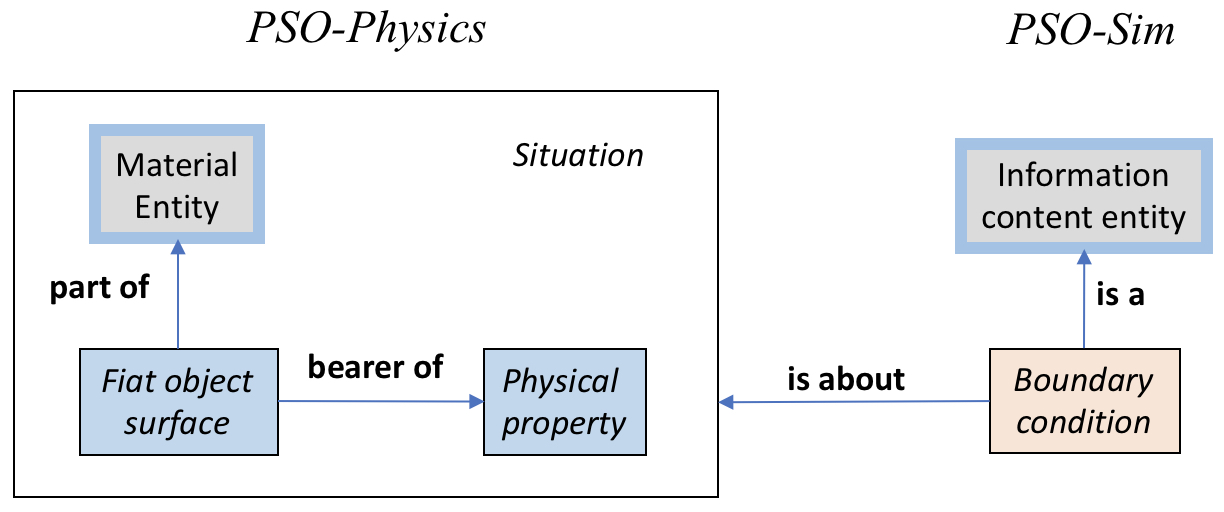}}
\subfigure[]{\includegraphics[width=0.58\linewidth]{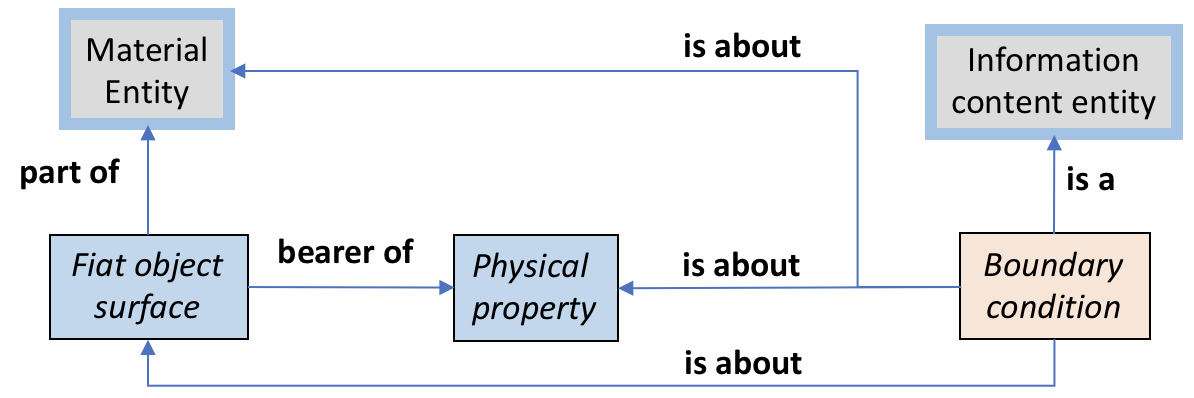}}
\caption{(a) Definition of a boundary condition involving a situation. (b) Simplified definition}
\end{center}
\end{figure}

%\paragraph*{{\small PSO: }\textit{Body condition.}}

%A body condition refers to a situation in which some material entity bears some physical property throughout the physical behavior of interest. Similar to a boundary condition, the physical properties vary depending on the types of body conditions and physics. Also, a simplified definition is used that does not require quantifying over relations. In simulation, body conditions are always applied on the entirety of domains, which represent the corresponding independent continuants in physical reality.

\paragraph*{{\small PSO: }\textit{Initial condition.}}

An initial condition refers to a situation in which some material entity bears some physical property at the onset of a physical behavior. Similar to a boundary condition, the physical properties vary depending on the types of the physical behavior involved, and a simplified definition that does not require quantifying over relations can be used. An initial condition is always applied on the entirety of a domain.

\paragraph*{{\small PSO: }\textit{Time step.}}

A time step is used in simulation to discretize the 1-D temporal region (or temporal interval) in which the physical behavior of interest is located. Therefore, it is about the temporal parts of a temporal interval that is partitioned into equal durations. 

%-------------------------------------------------

\subsection{Relation in PSO-Sim}

The following relation is used to relate entities between PSO-Sim and PSO-Physics.

\paragraph*{\textit{is about.}}

This relation is adopted from \citep{ceusters2015aboutness}, where it is used to model how a representational entity \textbf{is about} some other entity. In the PSO context, the relation is held between an instance of a PSO-Sim universal (ICE) and an instance of a PSO-Physics universal, which the former refers to provide input data for simulation. For example, a particular mesh file \textbf{is about} the shape of a car.

%\paragraph*{\textit{information part of.}}

%This relation is held between two instances of ICEs, and has been created to model how the former is part of the latter. For example, a person's first name is \textbf{information part of} that person's full name. In the PSO context, this relation is used to indicate that a simulation problem definition contains different ICE parts such as domains, boundary conditions, time steps, etc.

%=================================================

\section{Case study: Modeling physics problems as input to simulation solvers}
\label{case}

To demonstrate the value of PSO, a case study was conducted. The case study involved modeling a multi-physics engineering analysis problem using the terms of PSO, and mapping that model into the required input for two different physics solvers, namely FEniCS and NASTRAN. FEniCS is an open-source computing platform for solving PDEs \citep{langtangen2016solving}, applicable for general physics problems. NASTRAN, on the other hand, is a finite element analysis program that focuses on solving structural-dynamic-thermal problems in engineering \citep{nastran}. It has been integrated into a number of commercial CAE software applications. To demonstrate the breadth of the modeling capability, input data for three separate physics problems were generated, two of which were targeted for both FEniCS and NASTRAN while one was targeted only for FEniCS (due to the fact that NASTRAN cannot handle that problem type). In addition, the case study shows how PSO aids in the consistent modeling of the problem and the reuse of data across multiple simulation solvers.

%-------------------------------------------------

\subsection{Description of an engineering analysis problem}

\begin{figure}[H]
\begin{center}
\includegraphics[width=0.68\linewidth]{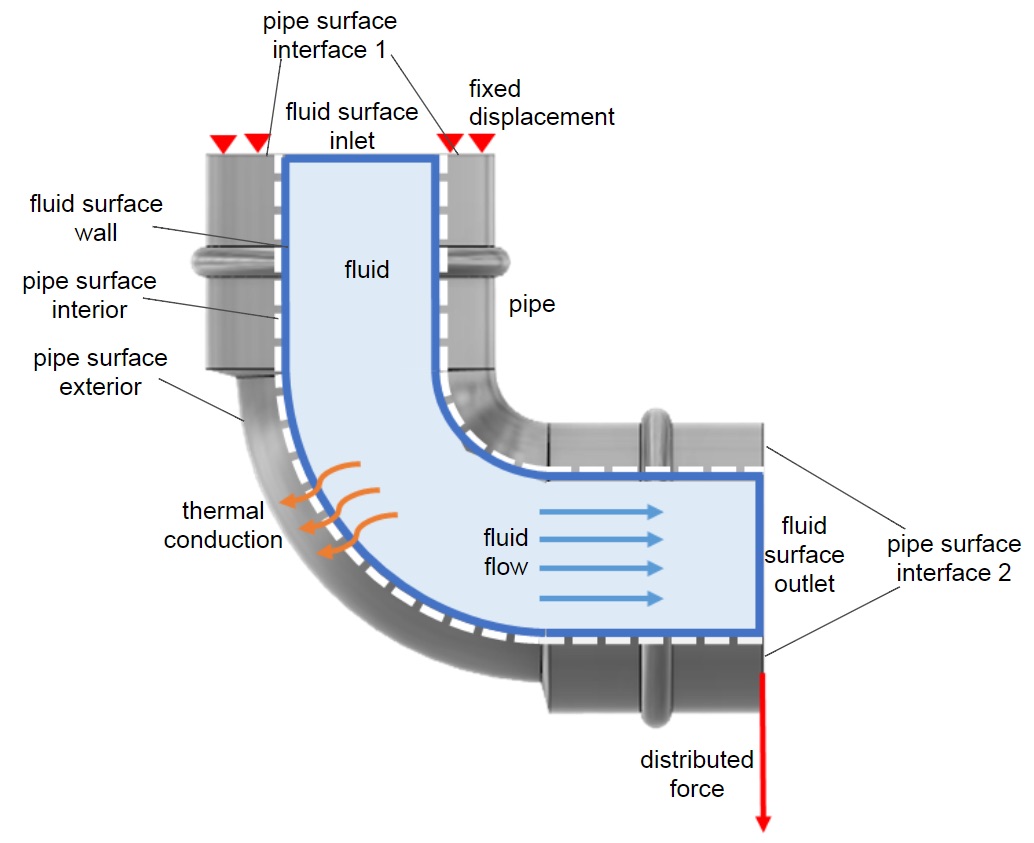}
\caption{Depiction of the example problem.}
\label{fig5}
\end{center}
\end{figure}

The problem chosen involved analyzing the physical behaviors of a simple pipe elbow with some fluid flowing through the hole in the pipe. The pipe is made of cast iron while the fluid is 5W-30 liquid oil. The pipe has its both ends connected to some other objects, where one end is assumed to be fixed to one object and the other end bears the weight of the other object. The fluid is hot and therefore transfers heat to the pipe via conduction. Figure~\ref{fig5} depicts the example problem.

\subsection{Modeling the problem using PSO}

The problem was modeled by identifying all the required information as instances of PSO-Physics, shown in Table 1, and relations between those instances, shown in Table 2. While the relations identified do not explicitly appear in the translation examples, they are necessary to ensure the correct association of individual entities when they are translated from one solver to another. For example, relations would ensure that ``density of cast iron" is associated with ``cast iron", not ``5W-30". Also, relations are essential for reasoning with the data model to check its consistency. For example, if an assertion such as ``cast iron \textbf{made of} pipe'', one could automatically identify that such assertion is incorrect compared to the axiom defined for the \textbf{made of} relation. 

Some of the PSO-Physics categories were extended with more specific categories to curate the data instances and aid in mapping them to solver input data. In actual applications, certain instances such as those of physical and material property types could be assigned with data values, e.g., ``displacement of PSO-1 \textbf{has value} 50".

\begin{table}[H]
\tbl{Instances of the PSO model of the example problem\\}{
  \begin{tabular}{l l p{7.5cm}}
  PSO Class & Extended Subclass & Instances \\\hline\toprule
  \emph{Process} & & fluid flowing through pipe \\\midrule
  \emph{Physical behavior} & \emph{Structural behavior} & structural behavior of pipe \\
   & \emph{Fluid behavior} & flow behavior of fluid \\
   & \emph{Thermal behavior} & thermal behavior of pipe \\\midrule
  \emph{1-D temporal region} & & fluid flow duration \\\midrule
  \emph{Material substance} & \emph{Cast Iron} & cast iron \\
   & \emph{Oil} & 5W-30 oil \\\midrule
  \emph{Object} & & pipe, (a portion of) fluid \\\midrule
  \emph{Object aggregate} & & pipe and fluid assembly \\\midrule
  \emph{Fiat object surface} & & pipe surface interface 1 (PSI-1), pipe surface interface 2 (PSI-2), pipe surface exterior (PSE), pipe surface interior (PSIr), fluid surface inlet (FSI), fluid surface outlet (FSO), fluid surface wall (FSW) \\\midrule
  \emph{Site} & & pipe hole \\\midrule
  \emph{Shape} & & shape of pipe, shape of fluid, shape of pipe hole \\\midrule
  \emph{Physical property} & \emph{Displacement} & displacement of PSI-1 \\
   & \emph{Body force} &  body force throughout pipe, body force throughout fluid \\
   & \emph{Distributed force} & distributed force at PSI-2 \\
   & \emph{Pressure} & pressure at FSI, pressure at FSO \\
   & \emph{Velocity} & velocity at FSW \\
   & \emph{Temperature} & initial temperature of pipe, temperature at PSE \\ 
   & \emph{Temperature flux} & heat source throughout pipe, temperature flux at PSIr \\\midrule
  \emph{Material property} & \emph{Density} & density of cast iron, density of 5W-30 oil \\
   & \emph{Elastic modulus} & elastic modulus of cast iron \\
   & \emph{Shear modulus} & shear modulus of cast iron \\
   & \emph{Specific heat capacity} & specific heat capacity of cast iron \\
   & \emph{Thermal conductivity} & thermal conductivity of cast iron \\
   & \emph{Viscosity} & viscosity of 5W-30 oil \\\midrule
  \emph{Contact} & & contact between PSIr and FSW \\\midrule
  \end{tabular}}
\end{table}

\subsection{Mapping to input data for FEniCS and NASTRAN}

Here, the example problem modeled is mapped to three separate physics problems to be solved. For the first two problems, which require linear static and thermal simulations, respectively, input data for both FEniCS and NASTRAN are identified. For the last problem, which requires fluid simulation, input data for only FEniCS are presented since the simulation cannot be performed with NASTRAN. The required input information for FEniCS is identified based on the tutorial examples provided in FEniCS's documentation \citep{langtangen2016solving}, adjusted to solve the current case study problem where necessary. The source code for each tutorial example is available on-line and cited below. For NASTRAN, the required information was identified based on the review of published user guides \citep{nx, thermal}.

\begin{table}[H]
\tbl{Relations of the PSO model of the example problem\\}{
  \begin{tabular}{l p{3.5cm} l}
  Instance 1 & PSO relation & Instance 2 \\\hline\toprule
  fluid flowing through pipe & \textbf{occupies t-region} & fluid flow duration \\
  fluid flowing through pipe & \textbf{has participant} & pipe and fluid assembly \\
  structural behavior of pipe & \textbf{process profile of} & fluid flowing through pipe \\
  flow behavior of fluid & \textbf{process profile of} & fluid flowing through pipe \\
  thermal behavior of pipe & \textbf{process profile of} & fluid flowing through pipe \\
  pipe & \textbf{continuant part of} & pipe and fluid assembly \\
  fluid & \textbf{continuant part of} & pipe and fluid assembly \\
  pipe hole & \textbf{continuant part of} & pipe \\
  pipe surface interface 1 & \textbf{continuant part of} & pipe surface exterior \\
  pipe surface interface 2 & \textbf{continuant part of} & pipe surface exterior \\
  pipe surface exterior & \textbf{continuant part of} & pipe \\
  pipe surface interior & \textbf{continuant part of} & pipe \\
  fluid surface inlet & \textbf{continuant part of} & fluid \\
  fluid surface outlet & \textbf{continuant part of} & fluid \\
  fluid surface wall & \textbf{continuant part of} & fluid \\
  fluid & \textbf{located in} & pipe hole \\
  pipe & \textbf{made of} & gray cast iron \\
  fluid & \textbf{made of} & 5W-30 oil \\
  shape of pipe & \textbf{s-depends on} & pipe \\
  shape of fluid & \textbf{s-depends on} & fluid \\
  shape of pipe hole & \textbf{s-depends on} & pipe hole \\
  displacement of PSI-1 & \textbf{s-depends on} & pipe surface interface 1 \\
  distributed force at PSI-2 & \textbf{s-depends on} & pipe surface interface 2 \\
  body force throughout pipe & \textbf{s-depends on} & pipe \\
  body force throughout fluid & \textbf{s-depends on} & fluid \\
  initial temperature of pipe & \textbf{s-depends on} & pipe \\
  heat source throughout pipe & \textbf{s-depends on} & pipe \\
  pressure at FSI & \textbf{s-depends on} & fluid surface inlet \\
  pressure at FSO & \textbf{s-depends on} & fluid surface outlet \\
  velocity at FSW & \textbf{s-depends on} & fluid surface wall \\
  temperature at PSE & \textbf{s-depends on} & pipe surface exterior \\  
  temperature flux at PSIr & \textbf{s-depends on} & pipe surface interior \\
  density of cast iron & \textbf{s-depends on} & cast iron \\
  elastic modulus of cast iron & \textbf{s-depends on} & cast iron \\
  shear modulus of cast iron & \textbf{s-depends on} & cast iron \\
  specific heat capacity of cast iron & \textbf{s-depends on} & cast iron \\
  thermal conductivity of cast iron & \textbf{s-depends on} & cast iron \\
  density of 5W-30 oil & \textbf{s-depends on} & 5W-30 oil \\
  viscosity of 5W-30 oil & \textbf{s-depends on} & 5W-30 oil \\
  contact btw. PSIr and FSW & \textbf{s-depends on} & pipe surface interior \\
  contact btw. PSIr and FSW & \textbf{s-depends on} & fluid surface wall \\
  \midrule
  \end{tabular}}
\end{table}

%-------------------------------------------------

\subsubsection{Linear elastic simulation}

The problem was mapped to the input data for the linear elastic tutorial example\footnote{https://github.com/hplgit/fenics-tutorial/blob/master/pub/python/vol1/ft06\_elasticity.py} in FEniCS and corresponding data elements in NASTRAN. Linear elastic simulation involves the governing equations (5) and (6) presented in Section 3.2, where $u$ represents the displacement of the pipe.
%\begin{gather} 
%\nabla\cdot\sigma = f \text{ in } \Omega \\
%\sigma = \lambda\,\text{tr}\,(\varepsilon) I + 2\mu\varepsilon \\
%\varepsilon = \frac{1}{2}\left(\nabla u + (\nabla u)^{\top}\right)
%\end{gather}

Specific boundary conditions for the case study problem are:
\begin{gather} 
u = \delta = 0 \text{ on } \partial\Omega_{D,\text{fixed}} \\
\mathbf n \cdot \sigma(u) = P \text{ on } \partial\Omega_{N,\text{pressure}}
%\dfrac{d\sigma(u)}{dn} = P \text{ on } \partial\Omega_{N,\text{pressure}}
\end{gather}

Table 3 shows how the problem modeled with PSO can be mapped to the corresponding elements of these equations and the required input data for FEniCS and NASTRAN. The top portion of the table contains PSO-Physics classes and their instances, which can be directly used as input to both FEniCS and NASTRAN once the values have been specified. In contrast, the bottom portion contains PSO-Sim classes and their newly created instances, which refer to the existing instances of PSO-Physics. Such instances are re-instantiated for different solvers because each solver employs its own way of representing mesh or defining boundary conditions, for example.\\

\begin{table}[H]
\tbl{The top/bottom portion of the table shows the mapping between the problem modeled with PSO-Physics/PSO-Sim and the corresponding data items that would be used/instantiated as input to FEniCS and NASTRAN for linear elastic simulation\\}{
  \begin{tabular}{p{2.7cm} p{4.4cm} p{1.7cm} p{1.3cm} p{2cm}}
  PSO Class & PSO Instance & Equation term & FEniCS data item & NASTRAN data item \\\hline\midrule
  \emph{Elastic modulus} & elastic modulus of cast iron & $\lambda$ in Eq (6) & ``lambda'' & MAT1.E\\
  \emph{Shear modulus} & shear modulus of cast iron & $\mu$ in Eq (6) & ``mu'' & MAT1.G\\
  \emph{Body force} & body force throughout pipe & $f$ in Eq (5) & ``f'' & GRAV.A\\
  \emph{Distributed force} & distributed force at PSI-2 & $P$ in Eq (11) & ``T''$^{\rm a}$ & PLOAD.P\\
  \emph{Displacement} & displacement of PSI-1 & $\delta$ in Eq (10) & ``d'' & SPC.D\\
  \midrule
  \emph{Mesh} & mesh that \textbf{is about} shape of pipe & $\Omega$ in Eq (5) & ``mesh'' & GRID+CHEXA\\
  \emph{Boundary condition (displacement)} & boundary condition that \textbf{is about} (pipe, pipe surface interface 1, displacement of PSI-1) & Eq (10) & ``bc'' & SPC\\
  \midrule
  \end{tabular}}
  \tabnote{$^{\rm a}$In FEniCS, Neumann boundary conditions are not explicitly specified but included as part of the variational form of the governing equation. Hence, only the physical properties involved need to be specified.}
\end{table}

\subsubsection{Heat transfer simulation}

Next, the case study problem was mapped to the input data for the heat transfer tutorial example\footnote{https://github.com/hplgit/fenics-tutorial/blob/master/pub/python/vol1/ft04\_heat\_gaussian.py} in FEniCS and corresponding data elements in NASTRAN. Heat transfer simulation involves solving the governing equation (1) in Section 3.1, over domain $\Omega$ and time $0<t<T$, for $u$ that represents the temperature of the pipe. 

Specific boundary conditions for the case study problem are:
\begin{gather} 
u = u_D \text{ on } \partial\Omega_{D,\text{exterior}} \\
\kappa \dfrac{du}{dn} = g \text{ on } \partial\Omega_{N,\text{interior}}
\end{gather}
and an initial condition is posed with the same equation as (4).

Table 4 shows how the problem modeled with PSO can be mapped to the corresponding elements of these equations and the required input data for FEniCS and NASTRAN. Again, the top and bottom portions of the table are divided based on the distinction between PSO-Physics and PSO-Sim instances, where the top portion of data can be used for both FEniCS and NASTRAN.\\

\begin{table}[H]
\tbl{The top/bottom portion of the table shows the mapping between the problem modeled with PSO-Physics/PSO-Sim and the corresponding data items that would be used/instantiated as input to FEniCS and NASTRAN for heat transfer simulation \\}{
  \begin{tabular}{p{2.8cm} p{4.3cm} p{1.8cm} p{1.3cm} p{2cm}}
  PSO Class & PSO Instance & Equation term & FEniCS data item & NASTRAN data item \\\hline\midrule
  \emph{1-D temporal region} & fluid flow duration & $T$ in Eq (1) & ``T'' & TIME\\
  \emph{Density} & density of cast iron & $\rho$ in Eq (1) & ``rho''$^{\rm b}$ & MAT4.$\rho$\\
  \emph{Specific heat capacity} & specific heat capacity of cast iron & $c_p$ in Eq (1) & ``cp''$^{\rm b}$ & MAT4.CP(T)\\
  \emph{Thermal conductivity} & thermal conductivity of cast iron & $\kappa$ in Eq (1) & ``kappa''$^{\rm b}$ & MAT4.K(T)\\
  \emph{Temperature flux} & heat source throughout pipe & $f$ in Eq (1) & ``f'' & QVOL.Q0\\
  \emph{Temperature flux} & temperature flux at PSIr & $g$ in Eq (13) & ``g''$^{\rm a}$ & TEMPBC.TEMP\\
  \emph{Temperature} & temperature at PSE & $u_D$ in Eq (12) & ``u\_D'' & QHBDY.Q0\\
  \midrule
  \emph{Mesh} & mesh that \textbf{is about} shape of pipe & $\Omega$ in Eq (1) & ``mesh'' & GRID+CHEXA\\
  \emph{Boundary condition (temperature)} & boundary condition that \textbf{is about} (pipe, pipe surface interface 1, displacement of PSI-1) & Eq (12) & ``bc'' & TEMPBC\\
  \emph{Initial condition (temperature)} & initial condition that \textbf{is about} (pipe, pipe surface interface 1, displacement of PSI-1) & Eq (4) & ``u\_n'' & TEMP(INIT)\\
  \midrule
  \end{tabular}}
  \tabnote{$^{\rm a}$In FEniCS, Neumann boundary conditions are not explicitly specified but included as part of the variational form of the governing equation. Hence, only the physical properties involved need to be specified.}
    \tabnote{$^{\rm b}$In the tutorial example, instead of material properties, nondimensionalized parameters that depend on those material properties are used. Here, we use the material properties for clarity.}
\end{table}

\subsubsection{Fluid simulation}

Lastly, the case study problem was mapped to the input data to the fluid (Navier-Stokes) tutorial example\footnote{https://github.com/hplgit/fenics-tutorial/blob/master/pub/python/vol1/ft07\_navier\_stokes\_channel.py} for FEniCS. Fluid simulation involves solving the following governing equations in domain $\Omega$ and for time $0<t<T$, where $u$ represents the fluid velocity and $p$ represents the fluid pressure:
\begin{gather} 
\nabla \cdot u = 0 \\
\rho\left(\frac{\partial u}{\partial t} +
  u \cdot \nabla u\right) = \nabla\cdot\sigma(u, p) + f \\
\sigma(u, p) = 2\mu\epsilon(u) - pI
\end{gather}
with the following boundary conditions:
\begin{gather} 
u = u_\text{walls} = 0 \text{ on } \partial\Omega_{D,\text{walls}} \\
\dfrac{dp}{dn} = p_\text{inflow} \text{ on } \partial\Omega_{N,\text{inflow}} \\
\dfrac{dp}{dn} = p_\text{outflow} \text{ on } \partial\Omega_{N,\text{outflow}}
\end{gather}
and the following initial condition for the case study problem:
\begin{gather}
u = u_\circ \text{ at } t=0 \text{, in } \Omega
\end{gather}
Table 5 shows how the problem modeled with PSO can be mapped to the corresponding elements of these equations and the required input data for FEniCS. Again, the top and bottom portions of the table are divided based on the distinction between PSO-Physics and PSO-Sim instances.\\

\begin{table}[H]
\tbl{The top/bottom portion of the table shows the mapping between the problem modeled with PSO-Physics/PSO-Sim and the corresponding data items that would be used/instantiated as input to FEniCS for fluid simulation \\}{
  \begin{tabular}{p{2.6cm} p{5.4cm} p{2.4cm} p{1.6cm}}
  PSO Class & PSO Instance(s) & Equation term & FEniCS data item$^{\rm a}$ \\\hline
  \midrule
  \emph{1-D temporal region} & fluid flow duration & $T$ in Eq (14-16) & ``T'' \\
  \emph{Viscosity} & viscosity of 5W-30 oil & $\mu$ in Eq (16) & ``mu'' \\
  \emph{Density} & density of 5W-30 oil & $\rho$ in Eq (15) & ``rho'' \\
  \emph{Distributed force} & body force throughout fluid & $f$ in Eq (15) & ``f'' \\
  \emph{Velocity} & velocity at FSW & $u_\text{walls}$ in Eq (17) & ``u\_walls'' \\
  \emph{Pressure} & pressure at FSI & $p_\text{inflow}$ in Eq (18) & ``p\_inflow'' \\
  \emph{Pressure} & pressure at FSO & $p_\text{outflow}$ in Eq (19) & ``p\_outflow'' \\
  \midrule
  \emph{Mesh} & mesh that \textbf{is about} shape of fluid & $\Omega$ in Eq (14-16) & ``mesh'' \\
  \emph{Boundary condition (velocity)} & boundary condition that \textbf{is about} (fluid, fluid surface wall, velocity at FSW) & Eq (17) & ``bcu\_noslip'' \\
  \emph{Boundary condition (pressure)} & boundary condition that \textbf{is about} (fluid, fluid surface inlet, pressure at FSI) & Eq (18) & ``bcp\_inflow'' \\
  \emph{Boundary condition (pressure)} & boundary condition that \textbf{is about} (fluid, fluid surface outlet, pressure at FSO) & Eq (19) & ``bcp\_outflow'' \\
  \midrule
  \end{tabular}}
  \tabnote{$^{\rm a}$Initial condition is automatically initialized by FEniCS as $u_\circ=0$.}
\end{table}

\subsection{Discussion of mapping examples}

The case study demonstrates the capability of PSO in modeling a multi-physics problem, which in turn was used as the required input to the chosen simulation solvers. Specifically, the mapping examples illustrate how the data modeled as part of PSO-Physics (the top portions of data in Tables 3-5) can be directly reused across the two solvers, while the additional data that are part of PSO-Sim (the bottom portions of data in Table 3-5) must be re-instantiated for each solver. In addition, it has been shown that even for those data that are not directly reused, their categorical structure identified via PSO-Sim stays consistent across the solvers. In other words, there are one-to-one correspondences between the PSO-Sim entities used in each solver, although their exact instantiations for each solver might be different. This suggests that the isomorphism of the two solver's data models can be preserved via the ontology.

\subsection{Implementation details}

Although the focus of the current paper is not about the implementation of the proposed ontology, some implementation details are given here to help the readers understand how its application would work.

First, an extended version of PSO-Physics, as demonstrated in the current case study, would be used in conjunction with a user interface to guide the user to model some physical phenomenon of interest. This workflow could be similar to working on a systems modeling environment such as OpenModelica or Simulink. The result of this process is the data containing the instances of PSO-Physics and their relations, as shown in Tables 1 and 2. Subsequently, when a solver is invoked to simulate the physical phenomenon modeled, the input data for the solver could be generated by directly taking some of the PSO-Physics instances while instantiating new PSO-Sim instances that are specific for the solver, via additional user interaction. The transfer of data between the original model and the input data model can be performed based on mappings established between the two data models, via common PSO classes, as seen in Tables 3 to 5. When the user requests new simulation with another solver, the same process would be repeated in which case the PSO-Physics instances can be reused again. Note that the mapping and the data translation processes involved are implemented by the simulation application developers, not the engineers who are the users of the simulation application.

%=================================================

\section{Discussion}

PSO has been developed to clearly distinguish the terms used to model physical phenomena, PSO-Physics, and the terms used to represent information that is about the physical phenomena, PSO-Sim. The former terms can be used to identify the physical behavior, along with its participating entities, to be simulated. The latter terms refer to the former terms and define the information required for a simulation solver, e.g., a discretized representation of the object to be simulated.

This clear distinction between PSO-Physics and PSO-Sim addresses the research questions outlined in Introduction. Having BFO as its roots, which aims to represent reality as veridically as possible, PSO-Physics is also designed to represent the physical phenomena of interest as they exist, independent of any solver-specific interpretations. The information modeled with PSO-Physics can then be consistent, shared, and reused across different solvers, assuming that all the solvers are aiming to simulate the same physical phenomenon. For each specific solver, PSO-Sim can be used to model additional information, referring to the original information modeled with PSO-Physics, that is specific to each simulation solver.

The current work can be seen as an attempt to apply BFO and ontological realism in the physics and engineering domain. It has shown that how mathematical constructs, which are often considered as abstract concepts, can be introduced to the ontology while not being confused with physical entities. Our approach is to introduce such mathematical concepts as information content entities that refer to corresponding physical entities. Using this approach, an ontology can be extended to include various mathematical concepts that are essential in simulation or any other computational procedures, with the clear separation of those concepts from the models of physical phenomena. In doing so, the ontology does not impose any specific view of mathematics during the initial modeling of physical phenomena. For example, the initial problem model generated for the case study in Tables 1 and 2 did not include any PSO-Sim notions such as boundary conditions or initial conditions, which are mathematical concepts. Again, this allowed clear separation between the types of information that can be reused across different solvers versus those that must be re-instantiated.

At the same time, the current work identified an important limitation of using BFO and ontological realism. Simulation can often be used to analyze the expected behavior of an object that does not exist yet, e.g., during conceptual design when the intended artifact has not been created. Strict ontological realism does not allow such \emph{possible} objects to be identified as a physical entity \citep{arp2015building}, but perhaps as some type of an information content entity. However, we would still like to treat such possible objects as if they already exist in reality, so that it can be assumed to be participating in the physical phenomenon to be simulated. Although this is departure from the ontological realism principles, we are not using this workaround to introduce any fictitious entities or relations that would contradict the laws of physics. That is, the possible world we are considering still must conform to the constraints of physical reality that is reflected in the ontology. Otherwise, the simulation results of a nonconforming model would be useless in practice.

Our development approach can be considered as a combination of top-down and bottom-up approaches. It is top-down in the sense that an existing ontology, BFO, was extended to create sub-category terms that are relevant in describing physics entities for our purpose. By reusing the backbone categorization established in BFO, we minimize the risk of making fundamental ontological errors in PSO. Following a realist ontology, we also avoided polluting PSO-Physics with application-specific interpretations and focused on modeling physical phenomena as they are observed. %For example, PSO-Physics use physical dependencies between realizable motions instead of kinematic joints to describe relative motions between two objects. The former strategy allows description of any relative motions in the future, based on which new types of kinematic joints can be defined.   

At the same time, we embraced the specific perspective taken by most of the simulation solvers, which is based on classical mechanics and formulating boundary value problems involving partial differential equations that explain physical behaviors. This perspective sets the requirements for the types of entities that need to be represented with our ontology, and the perspective is concretized in PSO-Sim. As a good example, the requirement of a boundary condition for solving simulation problems identified the need for the notion of a \emph{situation} in the ontology, which was missing in BFO.

%It should be noted, however, that committing to a particular perspective is not equivalent to identifying some solver-specific information categories and attempting to abstract them into concepts that have some correspondence in reality. The latter practice is done in much of the existing commercial simulation software, e.g., Simulink has created \emph{wrapper} concepts around a set of mathematical expressions as blocks that can be composed to build a simulation model. This pure bottom-up approach ties a particular concept with a specific mathematical interpretation, e.g., a ``beam'' block is always represented as a specific beam equation that has been implemented in the software. If another method of simulating a ``beam'' is found, a new concept category would have to be created to wrap the corresponding equation(s). In contrast, our approach can incorporate different mathematical interpretations in the future by reflecting them in PSO-Sim while keeping PSO-Physics and the data modeled with PSO-Physics constant.

So far, PSO-Physics does not contain in-depth taxonomies for categories such as material substances, physical properties, or energy, etc. This is intentional as we would like to consider PSO-Physics as a semi-\emph{formal} ontology (formal in the sense that it is domain-independent). Much like upper ontologies, our hope is that one could take PSO and easily extend it for their specific applications. If multiple ontologies with in-depth taxonomies are all developed as extensions of PSO, they can be more easily integrated based on the common root terms that they share, compared to the scenario in which all of them are developed from scratch. In that regard, PSO can be considered as a middle-level reference ontology that can serve as a bridge between more application-specific ontologies for physics-based simulation, and also to support their alignment to BFO if desired.

%=================================================
\section{Conclusions and future work}

The current work presented Physics-based Simulation Ontology (PSO), developed to support the modeling and reuse of data for physics-based simulation. PSO is intended to provide the common view through which physical phenomena can be veridically represented, at first independent of solver-specific interpretations, and then translated into the specific input data required for different solvers. This framework allows the former set of information modeled to be reused across different solvers.

PSO was extended from BFO, hence embracing realism as much as possible and adopting the ontological framework established by BFO. This approach led to the primary upper-level distinction of categories in PSO, namely between PSO-Physics and PSO-Sim. In addition, PSO terms were defined as specializations of BFO terms, which was a more guided process than defining these terms from scratch. In the future, it would be easier to integrate PSO with other ontologies developed with BFO as their basis, as they would share the common root terms as PSO. In these regards, the current work has valued the benefit of developing a new ontology by extending an existing upper ontology. In addition, our work has attempted applying BFO in the physics and engineering domain, identifying some of its limitations and workarounds developed to address them.

To complete the proposed ontological approach, other reference ontologies related to engineering should be integrated with PSO. For example, the material substance and material property categories in PSO could be further extended with certain existing materials ontologies developed. Interestingly, there are materials ontologies developed using BFO, e.g., \cite{premkumar2014semantic} and \cite{furini2016development}, which could be more easily combined with PSO as they would be sharing common root terms. Other important reference ontologies to consider are various ontologies of physical quantities and units \citep{haasquantities,gkoutos2012units,lefort2005ontology}.

Another significant aspect of future work is to develop supporting technologies that can aid in transformation between different data formats. Although our case study has shown semantic mapping between FEniCS and NASTRAN via PSO classes, the actual data may persist in different data formats. Hence, transformation methods between various data formats and the ontology data format are required, e.g., XML to OWL \citep{bohring2005mapping}, JSON to OWL, \citep{wischenbart2013automatic, cheong2019translating}, and EXPRESS to OWL \citep{pauwels2016express}, etc..

Lastly, PSO should be published in a digital artifact form along with detailed documentation, which would facilitate in adoption of the ontology in practical scenarios and help validate the usefulness of PSO. Also, formal axioms should be presented so that the logical consistency and the inference capabilities of the ontology could be demonstrated. 

% Finally, PSO must be tested through its aggressive use. The ontology should be validated in scenarios that require coordination of multiple simulation solvers, ideally in a cloud-based open environment, e.g., as envisioned in \cite{ivezic2018}. During this effort, the categories of PSO that are consistent across different data models can be identified and preserved, while other categories should be open for revision and modification. However, such validation process can only be achieved if there is a community of researchers who is willing to adopt and try PSO as the reference ontology for physics-based simulation. To encourage such acceptance, the future work will include creating detailed documentation with formalized axioms and a concrete version of the ontology, and publishing them on an open-source repository.

%=================================================

\bibliography{ref}
\bibliographystyle{tfcad}

%=================================================

\appendix

\newpage
\section{Terms of PSO-Physics not presented in Section 4}
\label{appx:A}

\subsection{Occurrents}

\paragraph*{{\small BFO: }\textit{Process boundary.}}

A process boundary is ``an occurrent entity that is the instantaneous temporal boundary of a process'' \citep{arp2015building}. In PSO, a process boundary can be used to identify the beginning and the end states of a physical process to be simulated.

\paragraph*{{\small BFO: }\textit{0-D temporal region.}}

A zero-dimensional temporal region is equivalent to a temporal or time instant \citep{arp2015building}. It can be used to identify the moment a physical process starts or ends.

\subsection{Qualities}

\paragraph*{{\small PSO: }\textit{Shape.}}

A shape is the external form of both material entities and immaterial entities. \\

\begin{minipage}{0.925\textwidth}
{\small PSO: }\textit{shape} = def. a {\small BFO: }\textit{quality} that is an external form of an {\small BFO: }\textit{independent continuant}. \\
\end{minipage}

Examples include the cylindrical shape of a pipe, roundness of the end surface of a column, or any free-form shapes of material entities and sites. Shapes, which inhere in physical entities in reality, are typically represented with geometric models in CAD / CAE applications. The term \emph{geometric model} is defined in PSO-Sim.

\paragraph*{{\small PSO: }\textit{State of matter.}}

A state of matter specifies whether a material entity is in a solid, liquid, gas, or plasma form. \\

\begin{minipage}{0.925\textwidth}
{\small PSO: }\textit{state of matter} = def. a {\small BFO: }\textit{quality} that is one of the distinct forms that a {\small PSO: }\textit{material entity} can exist, such as as solid, liquid, gas, or plasma.
\end{minipage}

\subsection{Specifically dependent continuants - Realizable entities}

\paragraph*{{\small PSO: }\textit{Energy.}}

Energy is categorized as a type of realizable entity because it is the bearing material entity's disposition to do work. Kinetic energy is a disposition of a pendulum to do work via its movement, strain energy is a disposition of a deformed object to ``undeform", thermal energy is a disposition of a furnace to transfer heat to its surroundings, electrical energy is a disposition of a generator to supply electrons, and so on. A disposition is a type of a realizable entity in BFO. \\

\begin{minipage}{0.925\textwidth}
{\small PSO: }\textit{energy} = def. a {\small BFO: }\textit{disposition} of a {\small PSO: }\textit{material entity} to do work of different forms, e.g., mechanical, thermal, electrical, etc., via an unfolding physical process.
\end{minipage}

\paragraph*{{\small PSO: }\textit{Field.}}

A field is also treated as a realizable entity, because it is generated from the bearing material entity as a disposition to affect other material entities co-located with the field. For example, a gravitational field arises from a bearing object, e.g., Sun, and it has the potential to induce gravitational forces on other objects co-located with the field. Another prominent example of fields are electromagnetic fields, which are generated from an electrically charged object and induce electromagnetic forces on other objects co-located with the field. \\

\begin{minipage}{0.925\textwidth}
{\small PSO: }\textit{field} = def. a {\small BFO: }\textit{disposition} of a {\small PSO: }\textit{material entity} to induce forces on other {\small PSO: }\textit{material entities} co-located with the field. \\
\end{minipage}

Energy and fields are perhaps the most contentious categories in PSO. Both entities are well-defined using mathematics, but less clear to be identified in reality because they are not directly observable and can only be indirectly measured. At the same time, they would no longer exist if the originating material entity ceases to exist (e.g., the gravitational field of Sun would no longer exist if Sun disappeared); hence, our commitment to classify them as specifically dependent continuants. In addition, they both can be thought of as dispositions of the objects to affect other objects, hence the classification as a realizable entity (disposition).

\paragraph*{{\small PSO: }\textit{Realizable motion.}}

In PSO, the notion of a realizable motion of an object is introduced, defined as follows: \\

\begin{minipage}{0.925\textwidth}
{\small PSO: }\textit{realizable motion} = def. a {\small BFO: }\textit{disposition} of a {\small PSO: }\textit{material entity} to move in some particular manner through an unfolding process. \\
\end{minipage}

For example, a piston placed in a cylinder has a realizable motion of sliding in a particular direction. A realizable motion could also be identified as stationary as in the case of the cylinder. Realizable motions that depend on each other form \emph{kinematic joints} often used in engineering. For instance, the relative dependency between the realizable motion of the piston (sliding) and the realizable motion of the cylinder (stationary) can be identified as a sliding joint, which is a type of a kinematic joint. 

Typically, kinematic joints can be defined in CAD software (or in mathematics), in idealized forms, using reference geometric entities such as points or lines \citep{demoly2012mereotopological,gruhier2015formal}. Such definitions of kinematic joints are not applicable in PSO-Physics that focus on physics entities, but can certainly be used in PSO-Sim that can deal with abstract entities. In PSO-Physics, we have created realizable motions so that kinematic joints defined in PSO-Sim refer to them as the desired motions to be simulated.

\newpage

\end{document}